\documentclass[10pt,twocolumn,letterpaper]{article}

\usepackage{cvpr}
\usepackage{times}
\usepackage{helvet} 
\usepackage{courier}  
\usepackage{epsfig}
\usepackage{graphicx}
\usepackage{amsmath}
\usepackage{amssymb}
\usepackage{multirow}
\usepackage{mathrsfs}
\usepackage{amsfonts}
\usepackage[dvipsnames]{xcolor}
\usepackage{soul}
\usepackage[numbers,sort]{natbib}
\usepackage{authblk}

\usepackage{amsmath}
\usepackage{etoolbox}
\usepackage{bm}
\usepackage{amsthm}
\usepackage{mathtools}
\usepackage{algorithm}
\usepackage{algpseudocode}
\usepackage{xspace}

\newcommand{\MyMapTemplateNoPrefix}[3]{\expandafter#1\csname#3\endcsname{#2{#3}}}
\forcsvlist{\MyMapTemplateNoPrefix {\def} {\mathbf} } {0,1,a,b,c,d,e, f, g, h, i, j, k, l, m, n, o, p, q, r, u, v, w, x, y, z} 

\newcommand{\MyMapTemplatePrefix}[4]{\expandafter#1\csname#3#4\endcsname{#2{#4}}} 
\forcsvlist{\MyMapTemplatePrefix {\def} {\mathcal}{c}} {A,B,C,D,E,F,G,H,I,J,K,L,M,N,O,P,Q,R,S,T,U,V,W,X,Y,Z}  
\forcsvlist{\MyMapTemplatePrefix {\def} {\mathbf} {b}} {A,B,C,D,E,F,G,H,I,J,K,L,M,N,O,P,Q,R,S,T,U,V,W,X,Y,Z} 
\forcsvlist{\MyMapTemplatePrefix {\DeclareMathOperator} {} {} } {trace, argmax, argmin, MixUp, Shuffle, Concat}

\def\etal{\emph{et al.}\@\xspace}
\def\ie{\emph{i.e.}\@\xspace}

\usepackage[breaklinks=true,bookmarks=false]{hyperref}

\cvprfinalcopy 

\def\cvprPaperID{6544} 

\ifcvprfinal\fi
\begin{document}

\title{TransMatch: A Transfer-Learning Scheme for Semi-Supervised Few-Shot Learning}

\author[1]{Zhongjie Yu\thanks{This work was done during Zhongjie's internship at Futurewei Technologies.}}
\author[2]{Lin Chen\thanks{Corresponding author: Lin Chen. Email: gggchenlin@gmail.com}}
\author[2]{Zhongwei Cheng}
\author[3]{Jiebo Luo}
\affil[1]{University of Wisconsin-Madison}
\affil[2]{Futurewei Technologies}
\affil[3]{University of Rochester}

\maketitle
\thispagestyle{empty}

\begin{abstract}

The successful application of deep learning to many visual recognition tasks relies heavily on the availability of a large amount of labeled data which is usually expensive to obtain. The few-shot learning problem has attracted increasing attention from researchers for building a robust model upon only a few labeled samples. 
Most existing works tackle this problem under the meta-learning framework by mimicking the few-shot learning task with an episodic training strategy. 
In this paper, we propose a new transfer-learning framework for semi-supervised few-shot learning to fully utilize the auxiliary information from labeled base-class data and unlabeled novel-class data. The framework consists of three components: 1) pre-training a feature extractor on base-class data; 2) using the feature extractor to initialize the classifier weights for the novel classes; and 3) further updating the model with a semi-supervised learning method. Under the proposed framework, we develop a novel method for semi-supervised few-shot learning called {\em TransMatch} by instantiating the three components with Imprinting and MixMatch. 
Extensive experiments on two popular benchmark datasets for few-shot learning, CUB-200-2011 and miniImageNet, demonstrate that our proposed method can effectively utilize the auxiliary information from labeled base-class data and unlabeled novel-class data to significantly improve the accuracy of few-shot learning task.
\end{abstract}

\section{Introduction}
Deep learning methods have been making impressive progress in different areas of artificial intelligence in recent years. Nevertheless, most of the popular deep learning methods require a large amount of labeled data which is usually very expensive and time-consuming to collect. The straightforward adoption of deep learning methods with a limited amount of labeled data usually leads to overfitting. Therefore, the question of whether it is able to learn a robust model from only a limited amount of labeled data arises.  It is well-known that humans have the ability to learn from a single or very few labeled samples. This motivates recent research efforts on learning a novel concept from a single or a few examples, \ie,~ few-shot learning.

In the past couple of years, an increasing number of few-shot learning methods have been proposed. One family of work focuses on training the model under the {\it meta-learning} framework based on an episodic training strategy~\cite{vinyals2016matching}. In particular, a sequence of episodes are randomly sampled where each episode consists of a few samples in the base classes to mimic the test scenario where only a few labeled samples of the novel classes are available. The labeled samples in each episode are divided into supports and queries, where supports are used for building the classifier and queries are used for evaluating.
At the same time, another family of work focuses on how to learn a classifier for the novel classes with only few-shot examples by transferring the knowledge from a model pre-trained on large amount of data from the base classes \cite{qi2018low,qiao2018few}. This paradigm shares similarity with human behaviors, by transferring past experience to new tasks. We denote this family of methods as {\it transfer-learning} based methods. Our method is inspired by the latter family of work and aims to learn a good classifier for the novel classes of few-shot examples with the help of the pre-trained classifier on abundant data from base classes and auxiliary unlabeled data from novel classes.

\begin{figure*}[ht]
    \centering
    \includegraphics[width=0.99\textwidth]{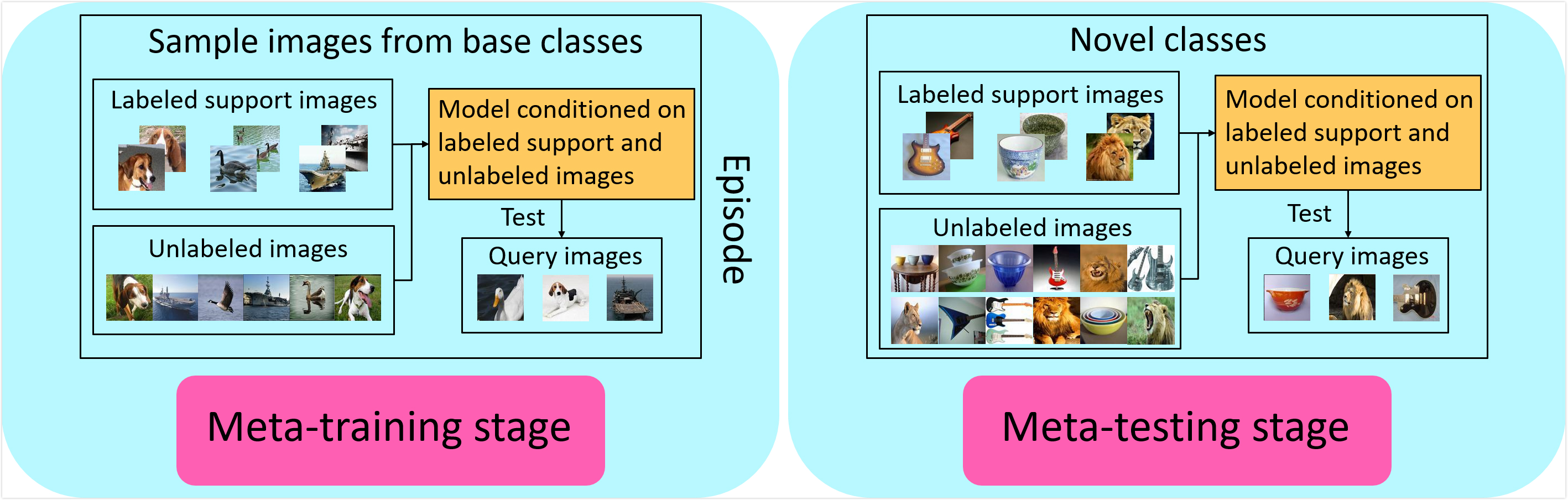}
    \caption{An overview of meta-learning based semi-supervised few-shot classification framework. Unlabeled images are required during training to allow the meta-learner learn how to leverage unlabeled images for classification.}
    \label{fig:conventional_framework}
\end{figure*}

We believe the sufficient and proper utilization of  extra information is crucial to the success of applying few-shot learning. Such extra information can exist in various forms, while in this work, we focus on leveraging extra information from the labeled base-class data and unlabeled novel-class data. These two types of information are usually easy to obtain. Many existing large-scale datasets for visual recognition tasks can be used for pre-training a model which can be later transferred to a new task. Meanwhile, it is also relatively easy to acquire a large amount of unlabeled data for a new task. Thus, a new paradigm called semi-supervised few-shot learning arises recently.


A representative work for semi-supervised few-shot learning \cite{ren2018meta} employed the meta-learning framework and enhanced the prototypical networks \cite{snell2017prototypical} to use unlabeled data. In each episode during meta-training, the unlabeled data for base classes was included to mimic the test scenario where the unlabeled data for novel classes would be available. Liu \etal~\cite{liu2018learning} proposed transductive propagation to incorporate the popular label propagation method to utilize the unlabeled data in episodic training. These works demonstrated that considering the unlabeled data helped to improve the accuracy of few-shot classification under the meta-learning framework. 

In this paper, we propose a new framework for semi-supervised few-shot learning to fully utilize the auxiliary information from labeled base-class data and unlabeled novel-class data. The flowchart of our proposed framework is shown in Fig.~\ref{fig:proposed_framework}, which consists of three components. 
We first train a model using the large amount of labeled data from the base classes, encoding the knowledge from base-class data into the pre-trained model. Then this pre-trained model is adopted as a feature extractor to generate the feature embeddings of the labeled few-shot examples from the novel classes, which can be directly used to imprint classifier weights for the novel classes or as the initialization of classifier weights for further fine-tuning, following the transfer-learning framework \cite{qi2018low}. Different from meta-learning, unlabeled images are no longer needed during pre-training on base classes, and could be directly utilized upon this imprinted classifier with state-of-the-art semi-supervised method such as MixMatch \cite{berthelot2019mixmatch}.  To the best of our knowledge, this is the first work of semi-supervised few-shot learning under the transfer-learning framework in contrast to the meta-learning framework. 

In summary, the contributions of our work are:
\begin{enumerate}
    \item We propose a new transfer-learning framework for semi-supervised few-shot learning, which can fully utilize the auxiliary information from labeled base-class data and unlabeled novel-class data.
    \item We develop a new method called {\em TransMatch} under the proposed framework. TransMatch integrates the advantages of transfer-learning based few-shot learning methods and semi-supervised learning methods, 
    and is different from the previous work on meta-learning based methods.
    \item We conduct extensive experiments on two popular benchmark datasets for few-shot learning to demonstrate that our method can effectively leverage unlabeled data in few-shot learning and achieve new state-of-the-art results. 
\end{enumerate}

\section{Related Work}
In this section, we review the related work to our proposed transfer-learning based semi-supervised few-shot learning framework.

\subsection{Few-Shot Learning}
Few-shot learning has attracted increasing attention in recent years due to the high cost of collecting labeled data. Existing work can be roughly categorized into (i) meta-learning methods, and (ii) transfer-learning methods. 

\begin{figure*}[tbp]
    \centering
    \includegraphics[width=0.99\textwidth]{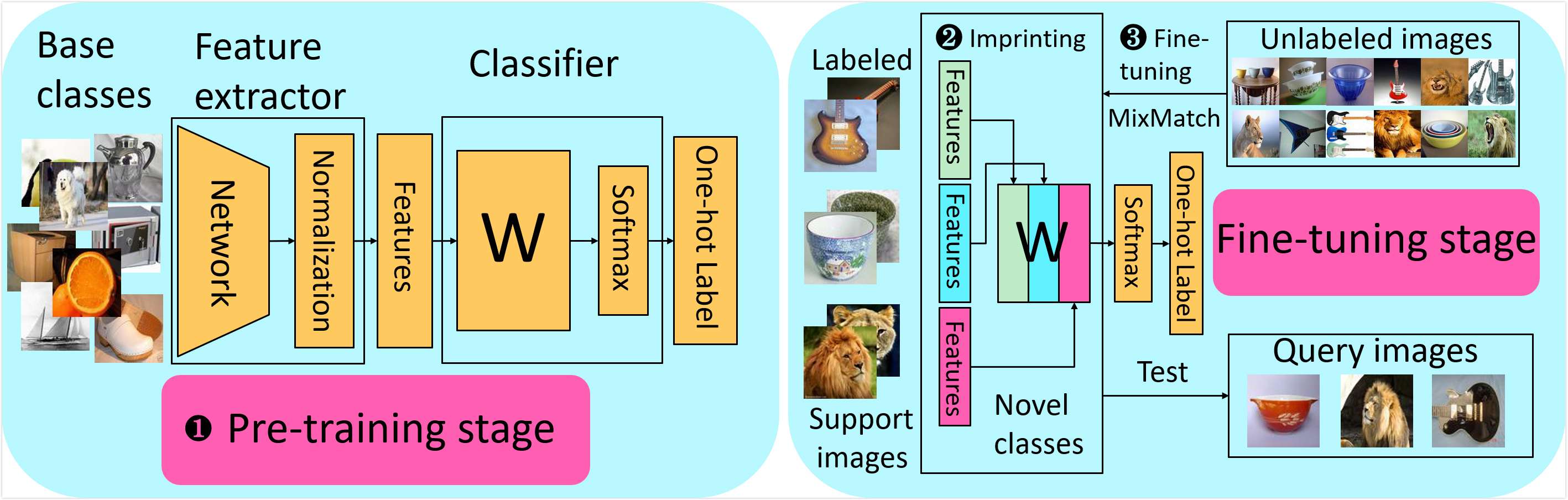}
    \caption{Our proposed framework of transfer-learning scheme for semi-supervised few-shot learning. We first pre-train a classifier from base-class images. Then use it as a feature extractor to initialize the weights for novel-class classifier. Finally, we further fine-tune the novel-class classifier with unlabeled images by semi-supervised learning method MixMatch.}
    \label{fig:proposed_framework}
\end{figure*}

\noindent{\bf Meta-learning based method:}\quad
Meta-learning based few-shot learning, also known as {\it learning to learn}, aims to learn a paradigm that can be adapted to recognize novel classes with only few-shot training examples. Meta-learning based methods usually consist of two stages: 1) meta-training; and 2) meta-testing. In the meta-training stage, a sequence of episodes are randomly sampled from the examples of base classes where each episode contains $K$ support examples and $Q$ query examples from $N$ classes, denoted as an $N$-way $K$-shot episode. In this way, the meta-training stage can mimic the few-shot testing stage where only a few examples per class are available. The meta-learning based methods can be further divided into two categories: a) metric-based methods; and b) optimization-based methods. 
    
{\bf  a) Metric-based methods} have been proposed in many existing  work~\cite{vinyals2016matching,snell2017prototypical,sung2018learning,oreshkin2018tadam,li2019revisiting}. These methods mainly focus on learning a good metric to measure the distance or similarity among support images and query images. For example, 
prototypical networks \cite{snell2017prototypical} calculated the distance of the prototype representations of each class between supports and queries.  
Relation Net \cite{sung2018learning} implemented a network to measure the relation similarities between the supports and queries.
Nearest Neighbour Neural Network \cite{li2019revisiting}  explored the nearest neighbors in local descriptors of feature embeddings.
    
{\bf b) Optimization-based methods} aim to design an optimization algorithm that can adapt the information during meta-training stage to the meta-testing stage.
Meta-LSTM \cite{ravi2016optimization} formulated the problem as an LSTM-based meta-learning algorithm to update the optimization algorithm in few-shot learning.
MAML \cite{finn2017model} learned an optimization method that can follow the fast gradient direction to rapidly learn the classifier for novel classes. 
LEO \cite{rusu2018meta} decoupled the gradient-based adaptation process with high-dimensional parameters to few-shot scenarios.
    
\noindent{\bf Transfer-learning based methods:} Transfer-learning based methods are different from meta-learning based methods, as they do not use the episodic training strategy. Instead, such methods can use conventional techniques to pre-train a model on the large amount of data from the base classes. The pre-trained model is then adapted to the few-shot learning task of recognizing novel classes. Qi \etal~\cite{qi2018low} proposed to imprint the classifier weights  of novel classes by the mean vectors of the feature embeddings of few-shot examples. Qiao \etal~\cite{qiao2018few} learned a mapping function from the activations (\ie,~feature embeddings) of novel class examples to classifier weights. Gidaris \etal~\cite{gidaris2018dynamic} proposed an attention module to dynamically predict the classifier weights for novel classes. Chen \etal~\cite{chen19closerfewshot} shown such transfer-learning based methods can achieve competitive performance as meta-learning based methods. 
Our proposed framework shares a similar idea with \cite{qi2018low} by pre-training a feature extractor and uses it to extract features for few-shot examples from novel classes which are used to imprint classifiers weights.

\subsection{Semi-Supervised Learning}\label{sec:related_work_ssl}
Semi-supervised learning focuses on developing algorithms to learn from unlabeled and labeled data. Existing work can be roughly categorized into (i) consistency regularization methods, and (ii) entropy minimization methods. 

\noindent{\bf Consistency regularization methods:} Consistency regularization methods mainly focus on adding noise and augmentation to images without changing their label distribution.
$\Pi$-Model \cite{Laine2017iclr} added a loss term to regularize the model by stochastic augmentation. Mean Teacher \cite{tarvainen2017mean} improved $\Pi$-Model by using the exponential moving average of parameters.
Virtual Adversarial Training (VAT) \cite{miyato2018virtual} regularized the model by  adding local perturbation on unlabeled data.

\noindent{\bf Entropy minimization methods:}\quad This family of methods focuses on giving low entropy for unlabeled data. It is initially proposed by \cite{grandvalet2005semi} which minimized conditional entropy of unlabeled data. Pseudo-Label \cite{lee2013pseudo} minimized the entropy directly by predicting the labels for unlabeled data and used this in cross-entropy, showing its good performance.

MixMatch \cite{berthelot2019mixmatch} united different kinds of consistency regularization and entropy minimization methods and achieved state-of-the-art performance by a large margin comparing with all the previous methods. It is a holistic method in semi-supervised learning and we would introduce briefly in Section \ref{sec:mixmatch}. Due to its good performance, we adopt MixMatch in our framework, and we also compared with using other mainstream semi-supervised learning methods in the experiments.
Semi-supervised learning methods are usually compared on small datasets \cite{oliver2018realistic,berthelot2019mixmatch,miyato2018virtual} where there is a small amount of labeled data. But the number of labeled images in typical semi-supervised learning is still greater than few-shot learning. The techniques for semi-supervised method may not be directly used for few-shot setting, which is also demonstrated in our experiments that naively applying MixMatch to few-shot learning may lead to poor performance especially in 1-shot and 2-shot.

\subsection{Semi-Supervised Few-Shot Learning} When there are few-shot examples for novel classes, it is straightforward to utilize extra unlabeled data to improve the learning. This leads to the family of semi-supervised few-shot learning methods (SSFSL). There are very few works in this direction. Ren \etal~\cite{ren2018meta} extended prototypical networks to incorporate unlabeled data by producing prototypes for the unlabeled data. Liu \etal~\cite{liu2018learning} constructed a graph between labeled and unlabeled data and utilize label propagation to obtain the labels of unlabeled data. 
Sun \etal~\cite{li2019learning} applied self-training by adding the confident prediction of unlabeled to the labeled training set in each round of optimization.

However, all existing semi-supervised few-shot learning methods are meta-learning based methods as in Fig.~\ref{fig:conventional_framework}. As shown in \cite{chen19closerfewshot}, transfer-learning  based method can achieve competitive performance compared with meta-learning based methods. This motivates our work. We need to emphasize that meta-learning based methods have shown their success to utilize unlabeled data by integrating unlabeled data in episodic training. However, this episodic training strategy is different from typical semi-supervised learning and it is not appropriate to combine them together directly. The techniques of leveraging unlabeled data in existing SSFSL methods are not state-of-the-art in semi-supervised areas and the more powerful and holistic methods like MixMatch would be difficult to integrate in meta-learning framework. Meanwhile, directly applying semi-supervised methods to utilize unlabeled data during test may lead to bad performance due to the extreme small number of labeled data.



\section{The Proposed Framework}
In this section, we introduce our proposed transfer-learning framework for semi-supervised few-shot learning. The flowchart is illustrated in Fig.~\ref{fig:proposed_framework}, which contains three modules: 1) pre-training a feature extractor on base-class data; 2) use the feature extractor to extract features from novel-class data and imprint novel-class classifier weights; and 3) further fine-tuning the model by semi-supervised learning method. Before elaborating the details of each module, let us first introduce our problem definition.

\noindent{\bf Problem definition:}
 We have a large-scale dataset $\cD_{base}$ containing many-shot labeled examples from each base class in $\cC_{base}$ and a small-scale dataset $\cD_{novel}$ of only few-shot labeled examples and many-shot unlabeled examples from each novel class in $\cC_{novel}$, where $\cC_{novel}$ is disjoint from $\cC_{base}$. The task of semi-supervised few-shot learning is to learn a robust classifier using both the few-shot labeled examples and many-shot unlabeled examples in $\cD_{novel}$ with the examples in $\cD_{base}$ as auxiliary data.  Usually in a conventional few-shot learning task, a small support set of $N$ classes with $K$ images per class is sampled from $\cD_{novel}$, leading to the $N$-way-$K$-shot problem. In semi-supervised few-shot learning, additional $U$ unlabeled images are sampled from each of the $N$ novel classes or distractor classes. 

\subsection{Part I: Pre-train Feature Extractor}
The first module of our framework, as shown in the left part of Fig.~\ref{fig:proposed_framework}, is a pre-training module, which relies on the many-shot examples from base classes, $\cD_{base}$, to train a base model which encodes as much as possible the information of $\cD_{base}$ and can be used in the later stage of few-shot learning as prior information, similar to human intelligence. This is different from conventional meta-learning based few-shot learning as shown in Fig.~\ref{fig:conventional_framework}, where an episodic training strategy is employed for base classes as well to mimic the few-shot scenario in the testing phase. 


\subsection{Part II: Classifier Weight Imprinting}
The weight imprinting method was proposed by \cite{qi2018low}, and has achieved impressive performance in the few-shot learning task as a representative of transfer-learning based few-shot learning method. Specifically, it directly sets the classifier weights by the mean feature vectors of the $N$-way-$K$-shot examples, where features are obtained by the model from the pre-training stage. For convenience, we denote the classifier on large scale base classes as $f(\x) = f^{base}(f^e(\x))$, where $\x$ is an input example, $f^e(\cdot)$ is the feature extractor and $f^{base}(\cdot)$ is the classifier. We have $f^e(\x) \in \cR^d$  and $f^{base}(\cdot) \in \cR^{|\cC_{base}|}$.

Given the $N$-way-$K$-shot examples from novel classes and let us denote them as $\cD_{novel}=\{\x_k^c |_{k=1...K,~c=1...N}\}$ with $\x_k^c$ as the $k$-th example in $c$-th class. We can use the feature extractor learned on base classes to extract features for the $N$-way-$K$-shot examples, denoted as $f^e(\x_k^c)$. Meanwhile, let us write the classifier for novel classes as $f^{novel}(\x) = \bW^\prime \x$, where $\bW = [\w_1, ..., \w_N] \in  \cR^{d\times N}$. Note that we omit the bias for simplicity. By normalizing the weight $\w_c$ and the feature vector $\x$ onto a unit ball, the aforementioned equation can be further simplified as 
\begin{eqnarray}
    f^{novel}(\x) = \left[\cos(\theta(\w_1, \x), ..., \cos(\theta(\w_N, \x))\right]^\prime,
\end{eqnarray}
where $\theta(\w_i, \x)$ denotes the angle between $\w_i$ and $\x$, and the classification for a given example $\x$ is based on computing the cosine similarity between every $\w_k$ and $\x$, and predict the label of $\x$ based on maximum similarity score.

In this sense, there is a duality between $\w_i$ and $\x$. Based on this observation, weight imprinting uses the mean feature vectors of the few-shot examples to imprint $\w_c$, \ie, by setting 
\begin{eqnarray}
    \w_c = \frac{1}{K}\sum_{k=1}^{K}f^e(\x_k^c).
\end{eqnarray}
The classification of an given example $\x$ can be also deemed as computing the mean of the similarities between $\x$ and all $K$-shot examples.

By imprinting the classifier weights with mean feature vectors of the few-shot examples, it  provides a better initialization of classifier weights to reduce the intra-class variations of features and benefits fine-tuning the new classifier for novel classes. Experimental results show that it can achieve good performance even without fine-tuning.

\subsection{Part III: Semi-Supervised Fine-tuning}\label{sec:mixmatch}
After we get the classifier which fully absorbs the information from base classes with a better initialization by imprinting, we fine-tune this classifier during test when there is unlabeled data. This fine-tuning process is the same as semi-supervised training. Any semi-supervised learning can be applied, and in this work we employed MixMatch \cite{berthelot2019mixmatch} not only because of its excellent performance in the semi-supervised learning task, but also because it is a holistic method to leverage unlabeled data in semi-supervised learning area. 

MixMatch combines multiple existing improvements from state-of-the-art semi-supervised learning methods which is discussed in Section \ref{sec:related_work_ssl}. In our setting, we denote $\cL = \{(\x_i, p_i)\}_{i=1}^B$ as a mini-batch of $B$ labeled examples with $p_i$ as the label, and $\cU=\{\x_u\}_{u=1}^U$ as a mini-batch of $U$ unlabeled examples. The imprinted classifier from Part II can be used to obtain estimated labels for the examples in $\cU$, \ie,~$f^{novel}(\x_u)$. We will omit the superscript $^{novel}$ for the ease of illustration when there is no confusion. For robustness, we augment each example $M$ times to get $M$ versions of each unlabeled data, \ie, $\{\x_{u,1}, ..., \x_{u,M}\}$, and use the mean prediction as the label estimation: $\bar{p}_u = \frac1M \sum_{i=1}^M f(\x_{u,i})$. The sharpen operation is used to enhance to prediction as $p_u = \bar{p}_u^{\frac1T}/\sum_{j=1}^N (\bar{p}_u)_j^{\frac1T}$, we set $T=0.5$ in the experiments.  The same data augmentation is also applied to labeled examples in $\cL$. Following \cite{berthelot2019mixmatch}, we concatenate $\cL$ and $\cU$ and shuffle the examples, \ie, $\cW = \Shuffle(\Concat(\cL, \cU))$, and then split this set into two new sets:
\begin{eqnarray}
    \cX_1' &=& \left\{ \MixUp \left(\cL_i,\cW_i\right) \;\;\;\;\;\;\;| i \in {1,\dots,|\cL|} \right\}, \nonumber\\
    \cX_2' &=& \left\{ \MixUp \left(\cU_i, \cW_{i+|\cL|}\right)\; | i \in {1,\dots,|\cU|}\right\} \nonumber,
\end{eqnarray}
where $\MixUp$ is defined as
\begin{flalign}
& \!\!\!\MixUp \left((\x_1, p_1),(\x_2, p_2)\right) \nonumber\\  
&  \;\;\;=\left((\lambda'\x_1+(1-\lambda')\x_2), (\lambda'p_1 + (1-\lambda')p_2)\right)
\end{flalign}
with $\lambda'= \max(\lambda, 1-\lambda)$. The parameter $\lambda$ is randomly generated from a beta distribution $\text{Beta}(\alpha,\alpha)$. 
The objective function to minimize is defined as
\begin{equation}
    \ell =\ell_1 + \gamma \ell_2,
\end{equation}
where
\begin{equation}
    \ell_1 = -\frac{1}{|\cX_1'|}\sum_{(\x, p)\in \cX_1'} p \log( f(\x)),
\end{equation}
is cross-entropy loss, 
and 
\begin{equation}
    \ell_2=\frac{1}{N|\cX_2'|}\sum_{(\x, p) \in \cX_2'}\left\|p - f(\x)\right\|_2^2.
\end{equation}
is consistency regularization loss in \cite{sajjadi2016regularization}.
The details of our algorithm is summarized in Algorithm~\ref{algo:1}.

\begin{algorithm}[h]
\begin{algorithmic}[1]
\Require{An auxiliary dataset $\cD_{base}$ with examples from $\cC_{base}$ (base classes),
$N$-way-$K$-shot dataset $\cD_l \!\!=\!\! \{\x_{nk},p|n\!=\!1,\cdots,N; k\!=\!1,\cdots,K\}$ with $p \in \cC_{novel}$ (novel classes), and $\cD_u = \{\x_u | u = 1,\cdots,U\}$}
\Ensure{$N$-way-$K$-shot classifier $f^{novel}$ for $\cD_l$ }
\State{Pre-train a base network on all examples in $\cD_{base}$ and denote it as $f^{base}(f^e(\x))$;}
\State{  Apply the feature extractor $f^e(\x)$ to extract features on $\cD_l$, then use these features to imprint the weights of the novel classifier $f^{novel}$;  }
\State{Apply semi-supervised learning method, MixMatch, to update the novel classifier $f^{novel}$ with both $\cD_l$ and $\cD_u$;}
\end{algorithmic}
\caption{Algorithm for our proposed TransMatch}\label{algo:1}
\end{algorithm}

\section{Experiments}
In this section, we evaluate our proposed TransMatch and compare with state-of-the-art few-shot learning methods on two popular benchmark datasets for few-shot learning, including miniImageNet and CUB-200-2011. 

\subsection{Experiments on miniImageNet}

\noindent{\bf Dataset configuration:}\quad The miniImageNet dataset was originally proposed by \cite{vinyals2016matching}. It has been widely used for evaluating few-shot learning methods. It consists of 60,000 color images from 100 classes with 600 examples per class, which is a simplified version of ILSVRC 2015 by \cite{deng2009imagenet}. We follow the split given by \cite{ravi2016optimization} consisting of 64 base classes, 16 validation classes and 20 novel classes. 
We randomly select $K$ (\emph{resp.} $U$) examples from each novel class as the few-shot labeled (\emph{unlabeled}) examples, and $Q$ images from the rest as the test examples. In the experiments, we set $N = 5$, $K = \{1, 5\}$, $Q = 15$ and study the effect of using different values of $U$. We repeat the test experiments $600$ times and report the mean accuracy with the 95\% confidence interval.

\noindent{\bf Compared methods:}\quad
The miniImageNet dataset has been widely used for evaluating the performance of few-shot learning methods, and is a good benchmark to compare state-of-the-art methods. In particular, we compare with several conventional few-shot learning methods, as well as state-of-the-art semi-supervised few-shot learning methods including the semi-supervised extension to Prototypical Networks by \cite{ren2018meta} (Soft k-Means,   Soft k-Means+Cluster,  Masked Soft k-Means), and  TPN-semi in \cite{liu2018learning}. We also re-implement Soft k-Means, Soft k-Means+Cluster, Masked Soft k-Means with the same backbone (\ie, WRN-28-10) as our method for fair comparison. As the area of semi-supervised few-shot learning has not been explored much yet, we also conduct extensive experiments to evaluate the performance of utilizing unlabeled data by our TransMatch under different few-shot settings.

\noindent{\bf Implementation details:}\quad
Following the work \cite{qiao2018few} for transfer-learning based method on miniImageNet, we use the wide residual network (\ie, WRN-28-10) \cite{zagoruyko2016wide} as the backbone for our base model $f^{base}$. We train it from scratch using the examples from the base classes. In particular, we first train a WRN-28-10 classification network on all examples from the 80 base and validation classes. 
We then replace the last layer of this network by a 256-d fully connected layer, followed by a L2 normalization layer and a 80-d classifier. We set the batch size to 128, and set learning-rate to 0.01 for the last two layers and 0.001 for all other layers.  We reduce the learning rate by 0.1 every 10 epochs and train for a total of 28 epochs. 

The base classifier $f^{base}$  is used as the feature extractor to generate feature vectors for the few-shot examples from novel classes. We use the few-shot labeled examples to fine-tune the base classifier to novel classes.  
We also augment each labeled image for 10 times by random transformation and use the mean features to imprint the weights for novel classifier. We use a batch size of 16, and set 64 batches as an epoch\footnote{We duplicate the labeled images dataset to make it larger, so that each batch may contain the same image multiple times.}. We set weight decay to 0.04, learning rate to 0.001, and use SGD optimizer with a momentum of 0.9. For the fine-tuning stage, we set the parameters of MixMatch as follows.  We set $M$ (the times for augmentation) to $2$, $T$ (the temperature for the label distribution) to $0.5$, $\gamma$ (the weight for regularization term) to $5$, $\alpha$ (the parameter in Beta distribution) to $0.75$. Meanwhile we use an exponential moving average for model parameters when guessing labels. 
For 5-way-1-shot scenario, we fine-tune for 10 epochs when there are 20 or 50 unlabeled images, and 20 epochs when there are 100 or 200 unlabeled images. For 5-way-5-shot scenario, we fine-tune for 20 epochs when there are 20 and 50 unlabeled images, and 25 epochs when there are 100 and 200 unlabeled images. All the test results are based on 600 random experiments.

\begin{table*}[tbp]
    \centering
    \begin{tabular}{l|l|c|c}
    \hline
        Method & Type & 1-shot & 5-shot\\
        \hline
        Prototypical Net \hfill\cite{snell2017prototypical} & Meta, Metric &49.42$\pm$0.78   &  68.20$\pm$0.66       \\
         TADAM \hfill \cite{oreshkin2018tadam} & Meta, Metric  & 58.50$\pm$0.30 & 76.70$\pm$0.30\\
        MAML \hfill\cite{finn2017model}& Meta, Optimization & 48.70$\pm$1.84 &63.11$\pm$0.92\\
         SNAIL \hfill \cite{mishra2017simple} & Meta, Optimization & 55.71$\pm$0.99& 68.88$\pm$0.92 \\
        Activation Net \hfill\cite{qiao2018few}& Transfer-learning & 59.60$\pm$0.41& 73.74$\pm$0.19\\ 
        Imprinting \hfill\cite{qi2018low}& Transfer-learning & 58.68$\pm$0.81 & 76.06$\pm$0.59\\
        \hline
        Soft k-Means \hfill \cite{ren2018meta}&Semi, Meta-learning & 50.09$\pm$0.45 & 64.59$\pm$0.28\\
        Soft k-Means+Cluster \hfill\cite{ren2018meta}&Semi, Meta-learning &  49.03$\pm$0.24 & 63.08$\pm$0.18\\
        Masked Soft k-Means \hfill\cite{ren2018meta}& Semi, Meta-learning & 50.41$\pm$0.31& 64.39$\pm$0.24\\
        TPN-semi \hfill\cite{liu2018learning} & Semi, Meta-learning &  52.78$\pm$0.27 & 66.42$\pm$0.21\\
        Soft k-Means \hfill(Re-implement with WRN-28-10) & Semi, Meta-learning & 51.88$\pm$0.93 & 67.31$\pm$0.70\\
        Soft k-Means+Cluster \hfill (Re-implement with WRN-28-10) &Semi, Meta-learning & 50.47$\pm$0.86 & 64.14$\pm$0.65\\
        Masked Soft k-Means \hfill(Re-implement with WRN-28-10)& Semi, Meta-learning & 52.35$\pm$0.89& 67.67$\pm$0.65\\
        \hline
        {TransMatch} \hfill(100 unlabeled images per class) & Semi, Transfer-learning &   \textbf{63.02$\pm$1.07} & 81.19$\pm$0.59 \\
        {TransMatch} \hfill(200 unlabeled images per class) & Semi, Transfer-learning &  62.93$\pm$1.11 & \textbf{82.24$\pm$0.59} \\
        \hline
    \end{tabular}
    \caption{Accuracy (in \%) on miniImageNet with 95\% confidence interval. Best results are in bold.}
    \label{tab:result_miniimagenet}
\end{table*}

\noindent{\bf Results on miniImageNet:}\quad
The results are summarized in Table \ref{tab:result_miniimagenet}. It is not surprising that our method outperforms conventional few-shot learning methods without using unlabeled by a large margin, as shown in the top portion of Table \ref{tab:result_miniimagenet}. Our method also outperforms  state-of-the-art semi-supervised few-shot learning methods, which can be observed from the middle portion of Table \ref{tab:result_miniimagenet}. These results clearly show the superiority of our TransMatch as its effective utilization of information from unlabeled data.

\begin{table}[tbp]
    \centering
    \begin{tabular}{l|c|c|c}
    \hline
        Method & \# unlabeled & 1-shot & 5-shot\\ \hline
     Imprinting   & -----  & 58.68$\pm$0.81 &76.06$\pm$0.59  \\ 
     Imprinting+FT& 0 & 55.60$\pm$0.77 & 74.17$\pm$0.60 \\
     {TransMatch} & 20 & 58.43$\pm$0.93 & 76.43$\pm$0.61\textbf{}   \\
     {TransMatch} & 50 & 61.21$\pm$1.03   & 79.30$\pm$0.59   \\
     {TransMatch} & 100 & \textbf{63.02$\pm$1.07}  & 81.19$\pm$0.59  \\ 
     {TransMatch} & 200 & 62.93$\pm$1.11 & \textbf{82.24$\pm$0.59} \\
     \hline
    \end{tabular}
    \caption{Accuracy (in \%) with different number of unlabeled images on miniImageNet. Best results are in bold.}
    \label{tab:result_miniimage_diff_unlabeled}
\end{table}

\noindent{\bf Influence of unlabeled examples:}\quad
In Table \ref{tab:result_miniimage_diff_unlabeled}, we report the results using different numbers of unlabeled images. Note that Imprinting+FT stands for fine-tuning the imprinted classifier without unlabeled data. It is obvious that our TransMatch could achieve better performance with more unlabeled images. We also observe that the results begin to saturate after 100 unlabeled images for 1-shot setting. In general, the results show that our TransMatch can effecively utilize the unlabeled data.

\begin{table}[tbp]
    \centering
    \begin{tabular}{c|c|c|c}
    \hline
        \# shot & Method & Accuracy & Gain \\
        \hline

        \multirow{2}*{1-shot} & w/ Pseudo-Label &  57.01 $\pm$ 1.13 & \multirow{2}*{+6.01}\\
        ~ & w/ MixMatch &  {63.02 $\pm$ 1.07} & ~\\
        \hline
        \multirow{2}*{2-shot} &w/ Pseudo-Label &  70.07 $\pm$ 0.96 & \multirow{2}*{+2.29} \\
        ~ & w/ MixMatch & {72.36 $\pm$ 0.88} & ~\\
        \hline
        \multirow{2}*{3-shot} &w/ Pseudo-Label & 76.01 $\pm$ 0.81 & \multirow{2}*{+1.40}  \\
        ~ & w/ MixMatch & {77.41 $\pm$ 0.76} & ~ \\
        \hline
        \multirow{2}*{4-shot} &w/ Pseudo-Label & 78.35 $\pm$ 0.73 & \multirow{2}*{+1.39} \\
        ~ & w/ MixMatch & {79.74 $\pm$ 0.65} & ~ \\
        \hline
        \multirow{2}*{5-shot} &w/ Pseudo-Label & 80.00 $\pm$ 0.66 & \multirow{2}*{+1.19} \\
        ~ & w/ MixMatch & {81.19 $\pm$ 0.59} & ~\\
        \hline
    \end{tabular}
    \caption{Comparison of our method using different semi-supervised learning methods (\ie,~Pseudo-Label and MixMatch) in our framework both with 100 unlabeled images for 5-way classification on miniImageNet.}
    \label{tab:result_miniimagenet_diff_shot_pseudo}
\end{table}

\begin{figure}[h]
    \centering
    \includegraphics[width=0.45\textwidth]{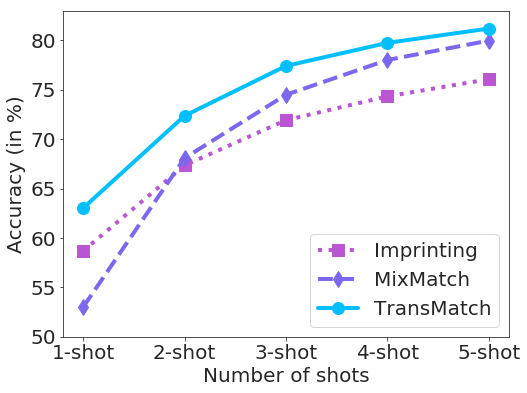}
    \caption{Comparison of Imprinting, MixMatch and our TransMatch both with 100 unlabeled images for 5-way classification with different number of shots on miniImageNet.}
    \label{fig:result_miniimagenet_diff_shot}
\end{figure}

\noindent{\bf Ablation study:}
We conduct an ablation study of our method without Imprinting or MixMatch. Without Imprinting, our method reduces to semi-supervised learning method, \ie, MixMatch (Note here the feature extractor is still already trained from base classes) and without MixMatch, our method reduces to Imprinting. The results are shown in Fig.~\ref{fig:result_miniimagenet_diff_shot}. It is clear that both MixMatch and Imprinting are worse than our TransMatch. 
The inferior performance of MixMatch to our TransMatch clearly shows that directly applying MixMatch to the few-shot setting cannot lead to good performance especially in 1-shot and 2-shot setting.
This is due to the lack of labeled data, which makes it hard to fine-tune the classifier during test when there is unlabeled data. However, our proposed TransMatch can obtain a good initialization by incorporating weight imprinting module. 

We also observe a larger gain by our TransMatch over MixMatch when using a smaller number of shots. The gain shown in Fig.~\ref{fig:result_miniimagenet_diff_shot} is $\{11.02, 4.28, 2.92, 1.73, 1.22\}$ in \{1, 2, 3, 4, 5\}-shot setting. This is reasonable and worth attention as fewer shots means fewer labeled examples, which makes fine-tuning more difficult. Therefore, the importance of weight imprinting to give the classifier good initial weights becomes more evident. 

\noindent{\bf Comparing different semi-supervised learning methods:}
In addition to MixMatch \cite{berthelot2019mixmatch}, in this section, we also compare with other semi-supervised learning methods (\ie, Pseudo-Label \cite{lee2013pseudo}) in order to understand the influence the semi-supervised learning module. The results, shown in Table~\ref{tab:result_miniimagenet_diff_shot_pseudo}, are consistent with our observations when using MixMatch as semi-supervised learning module. Since Pseudo-Label is worse than MixMatch, the overall performance of our method using Pseudo-Label is also worse than using MixMatch. 



\begin{table}[tbp]
    \centering
    \begin{tabular}{c|c|c|c}
    \hline
        Distractor & Method & 1-shot & 5-shot  \\
        \hline
        ----- & Imprinting & 58.68 $\pm$ 0.81 &  76.06 $\pm$ 0.59 \\
        \hline
        \multirow{2}*{1-class} & MixMatch & 50.14 $\pm$ 1.06 & 79.32 $\pm$ 0.63  \\
        ~ & TransMatch & \textbf{62.32 $\pm$ 1.04}  & \textbf{80.28 $\pm$ 0.62} \\
        \hline
        \multirow{2}*{2-class}  & MixMatch & 50.68 $\pm$ 1.15 & 78.07 $\pm$ 0.69\\

        ~ & TransMatch & \textbf{60.41 $\pm$ 1.02} & \textbf{79.48 $\pm$ 0.64} \\
        \hline
        \multirow{2}*{3-class}  & MixMatch &   49.48 $\pm$ 1.16  & 77.48 $\pm$ 0.66 \\

        ~ & TransMatch & \textbf{59.32 $\pm$ 1.10} & \textbf{79.29 $\pm$ 0.62} \\
        \hline
    \end{tabular}
    \caption{Accuracy (in \%) of MixMatch and our TransMatch with 100 unlabeled images from $\{1, 2, 3 \}$  {\it distractor} classes on  miniImageNet. Note that Imprinting does not use any unlabeled image.}
    \label{tab:result_miniimagenet_distractor}
\end{table}

\vspace{0.1in} 

\noindent{\bf Influence of distractor classes:}
In typical semi-supervised learning, unlabeled images come from the same classes for the labeled images. This may not reflect realistic situations in real-world application. So we also study the influence of distractor classes, and report the results of Imprinting, MixMatch, and our TransMatch when there are unlabeled images from various distractor classes. In our experiments, distractor classes are randomly chosen from the remaining classes which are disjoint with the novel classes during test. The results are shown in Table \ref{tab:result_miniimagenet_distractor}. We can observe that all the results for MixMatch degrade due to the distractor classes, while our TransMatch still outperforms Imprinting in all cases.

\subsection{Experiments on CUB-200-2011}
 
\noindent{\bf Dataset configuration:}\quad
The CUB-200-2011 dataset (CUB) is originally proposed by \cite{wah2011caltech} and contains 200 fine-grained classes of birds with 11,788 images in total (about 30  images per class for support images and 30 images per class for query images). We strictly follow the setup in \cite{qi2018low} to ensure a fair comparison. In particular, we use the standard train/test split provided by the dataset, and treat the first 100 classes as the base classes $\cC_{base}$  and the remaining 100 classes as the novel classes $\cC_{novel}$. Therefore, we have $N=100$. We use all the training examples from the base classes for large scale pre-training to obtain the base model $f^{base}$ and use the few-shot examples from the novel classes to train $f^{novel}$. In the experiment, we set $K$ to $\{1, 2, 5, 10, 20\}$ and use the rest images $\{29, 28, 25, 20, 10\}$ as unlabeled images for support images. All the remaining 30 images are still used for query images. 

\vspace{0.1in} 

\noindent{\bf Implementation details:}\quad
We are interested in performance of our TransMatch on the 100 novel classes, \ie, the {\it transfer-learning} setting in \cite{qi2018low}. In order to ensure fair comparison, we follow  \cite{qi2018low} and use Inception\_v1 as our network backbone. We set the dimension of  the fully connected embedding layer to 256, followed by an L2 normalization. 
We resize the input images to $256\times 256$ and then randomly crop to $224\times 224$.
During the large scale pre-training stage, we set the initial learning rate to 0.001 and a 10$\times$ multiplier for the embedding layer and classification layer. We reduce the learning rate by 0.1 after every 30 epochs, and train the model for a total of 90 epochs. 
During the fine-tuning stage, we set the number of batches to 64 for each epoch with a batch size of 64. By default, we set the weight decay to 0.0001, use a learning rate of 0.001, and train the model for 100 epochs. For the extreme case of 1-shot and 2-shot settings (100-way), we set the weight decay to 0.04, the learning rate to 0.0001 and early stopping at 10 epochs in order to avoid overfitting.

\noindent{\bf Results on CUB-200-2011:}
We follow \cite{qi2018low} to report the results of their proposed Imprinting, and Imprinting+FT. Then we evaluate the performance of our proposed TransMatch using different numbers of shots and unlabeled images. We compare TransMatch with Imprinting and MaxMatch in Table \ref{tab:result_cub}, and the results show our proposed TransMatch achieves the best result which demonstrates its effectiveness in utilizing auxiliary labeled base-class data and unlabeled novel-class data.  Table \ref{tab:result_cub_diff_unlabeled} shows the results of our TransMatch using different numbers of unlabeled images, and we can observe that better performance can be achieved with more unlabeled data. These results are similar to the results on miniImageNet dataset.

\begin{table}[tbp]
 \centering
 \begin{tabular}{l|r|r|r|r|r}
 \hline
     Model \hfill K= & 1 & 2 & 5 & 10 & 20 \\
     \hline
    Imprinting  & 26.08 & 34.13 & 43.34 & 48.91 & 52.94\\
    Imprinting+FT  & 26.59 & 34.33 & 49.39 & 61.65 & 70.07\\
    MixMatch  & 22.93  & 30.24 & 56.41  & 67.13 & 73.00 \\
    {TransMatch} \hfill & {\bf 28.02}  & \textbf{38.05} & \textbf{59.83} & \textbf{68.60} & \textbf{74.61} \\
    \hline
 \end{tabular}
 \caption{Accuracy (in \%) comparison  on CUB-200-2011. Best results are in bold.}
 \label{tab:result_cub}
 \end{table}
 
 
 \begin{table}[tbp]
    \centering
    \begin{tabular}{l|c|c|c}
    \hline
        Model & \# unlabeled & 5-shot & 10-shot\\
        \hline
     Imprinting \cite{qi2018low}  & -----    & 43.34 & 48.91 \\
     Imprinting+FT \cite{qi2018low} & 0   & 49.39 & 61.65\\
     TransMatch & 5 & 52.90    & 63.79\\
     TransMatch & 10 &  54.78   & 66.21 \\ 
     TransMatch & 15 &  56.86   & 67.71  \\
     TransMatch & 20 &  59.25  & 68.60\\
     \hline
    \end{tabular}
    \caption{Accuracy (in \%) comparison using different numbers of unlabeled images on CUB-200-2011.}\vspace{-0.2in}
    \label{tab:result_cub_diff_unlabeled}
\end{table}

\section{Conclusion}
While almost all existing semi-supervised few-shot learning methods are based on the meta-learning framework, we propose a new transfer-learning framework for semi-supervised few-shot learning to effectively explore the information from labeled base-class data and unlabeled novel-class data. We develop a new method under the proposed framework by incorporating the state-of-the-art semi-supervised method MixMatch and few-shot learning method Imprinting, leading to a new method called TransMatch. Extensive experiments on two popular few-shot learning datasets show that our proposed TransMatch achieves the state-of-the-art results, which demonstrate its effectiveness in utilizing both the labeled base-class data and unlabeled novel-class data.

{\small
\bibliographystyle{ieee_fullname}
\bibliography{arxiv}
}

\end{document}